\documentclass[11pt,a4paper]{article}
\usepackage[hyperref]{acl2018}
\usepackage{times}
\usepackage{latexsym}

\usepackage{url}
\usepackage{soul}
\usepackage{amsmath,array,graphicx}
\usepackage[export]{adjustbox}
\usepackage{caption}
\usepackage{subcaption}
\usepackage{multirow}
\usepackage{hyperref}
\usepackage{mathtools} 
\usepackage{graphicx} 
\usepackage{tabularx}
\usepackage{breakcites}
\usepackage{caption}

\usepackage{array}
\usepackage{makecell}
\usepackage{tabularx}

\usepackage{algorithm}
\usepackage{algorithmic}

\usepackage[inline, shortlabels]{enumitem}
\usepackage{float}

\aclfinalcopy 

\title{Sentiment Adaptive End-to-End Dialog Systems}

\author{Weiyan Shi \\
  {[}24{]}7.ai \\
  {\tt weiyan.shi@247.ai} \\\And
  Zhou Yu \\
  University of California, Davis  \\
  {\tt joyu@ucdavis.edu} \\}

\date{}

\begin{document}
\maketitle
\begin{abstract}
End-to-end learning framework is useful for building dialog systems for its simplicity in training and efficiency in model updating. However, current end-to-end approaches only consider user semantic inputs in learning and under-utilize other user information. Therefore, we propose to include user sentiment obtained through multimodal information (acoustic, dialogic and textual), in the end-to-end learning framework to make systems more user-adaptive and effective.
We incorporated user sentiment information in both supervised and reinforcement learning settings. In both settings, adding sentiment information reduced the dialog length and improved the task success rate on a bus information search task. This work is the first attempt to incorporate multimodal user information in the adaptive end-to-end dialog system training framework and attained state-of-the-art performance.

\end{abstract}

\section{Introduction}

Most of us have had frustrating experience and even expressed anger towards automated customer service systems. Unfortunately, none of the current commercial systems can detect user sentiment and let alone act upon it. Researchers have included user sentiment in rule-based systems \cite{acosta2009using, pittermann2010emotion}, where there are strictly-written rules that guide the system to react to user sentiment. Because traditional modular-based systems are harder to train, to update with new data and to debug errors, end-to-end trainable systems are more popular. However, no work has tried to incorporate sentiment information in the end-to-end trainable systems so far to create sentiment-adaptive systems that are easy to train. 
The ultimate evaluators of dialog systems are users. Therefore, we believe dialog system research should strive for better user satisfaction.
In this paper, we not only included user sentiment information as an additional context feature in an end-to-end supervised policy learning model, but also incorporated user sentiment information as an immediate reward in a reinforcement learning model. We believe that providing extra feedback from the user would guide the model to adapt to user behaviour and learn the optimal policy faster and better. 


There are three contributions in this work: 1) an audio dataset\footnote{The dataset is available \href{https://github.com/wyshi/sentiment-adaptive-dialog-systems}{\color{blue}{here}}.} 
with sentiment annotation (the annotators were given the complete dialog history); 2) an automatic sentiment detector that considers conversation history by using dialogic features, textual features and traditional acoustic features; and 3) end-to-end trainable dialog policies adaptive to user sentiment in both supervised and reinforcement learning settings. We believe such dialog systems with better user adaptation are beneficial in various domains, such as customer services, education, health care and entertainment. 

\section{Related Work}

Many studies in emotion recognition \cite{schuller2003hidden, nwe2003speech, bertero2016real} have used only acoustic features. But there has been  work on emotion detection in spoken dialog systems incorporating extra information as well \cite{Toward-Detecting-Emotions-in-Spoken-Dialogs, devillers2003emotion, liscombe2005using, burkhardt2009emotion, yu2017learning}. For example, 
\citet{liscombe2005using} explored features like users' dialog act, lexical context and discourse context of the previous turns. Our approach  considered 
accumulated dialogic features, such as total number of interruptions, to predict user sentiment along with acoustic and textual features.

The traditional method to build dialog system is to train modules such as language understanding component, dialog manager and language generator separately \cite{levin2000stochastic, williams2007partially, singh2002optimizing}. Recently, more and more work combines all the modules in an end-to-end training framework \cite{wen2016,li2017end,dhingra2016end,hybrid-code-networks-practical-efficient-end-end-dialog-control-supervised-reinforcement-learning-2,liu2017end}. Specifically related to our work, \citet{hybrid-code-networks-practical-efficient-end-end-dialog-control-supervised-reinforcement-learning-2} built a model, which combined the traditional rule-based system and the modern deep-learning-based system, with experts designing actions masks to regulate the neural model. Action masks are bit vectors indicating allowed system actions at certain dialog state. The end-to-end framework made dialog system training simpler and model updating easier.

Reinforcement learning (RL) is also popular in dialog system building \cite{zhao2016towards,liu2017iterative,li2016user}. A common practice is to simulate users. However, building a user simulator is not a trivial task. \citet{zhao2016towards} combines the strengths of reinforcement learning and supervised learning to accelerate the learning of a conversational game simulator. \citet{li2016user} provides a standard framework for building user simulators, which can be modified and generalized to different domains. \citet{liu2017iterative} describes a more advanced way to build simulators for both the user and the agent, and train both sides jointly for better performance.  We simulated user sentiment by sampling from real data and incorporated it as immediate rewards in RL, which is different from common practice of using task success as delayed rewards in RL training. 

Some previous module-based systems integrated user sentiment in dialog planning \cite{acosta2009using,acosta2011achieving,pittermann2010emotion}. They all integrated user sentiment in the dialog manager with manually defined rules to react to different user sentiment and showed that tracking sentiment is helpful in gaining rapport with users and creating interpersonal interaction in the dialog system. In this work, we include user sentiment into end-to-end dialog system training and make the dialog policy learn to choose dialog actions to react to different user sentiments automatically. We achieve this through integrating user sentiment into reinforcement reward design. Many previous RL studies used delayed rewards, mostly task success. However, delayed rewards make the converging speed slow, so some studies integrated estimated per-turn immediate reward. For example, \citet{ferreira2013expert} explored expert-based reward shaping in dialog management and \citet{ultes2017domain} proposed \textit{Interaction Quality (IQ)}, a less subjective variant of user satisfaction, as immediate reward in dialog training. However, both methods are not end-to-end trainable, and require manual input as prior, either in designing proper form of reward, or in annotating the \textit{IQ}.  Our approach is different as we detect the multimodal user sentiment on the fly and does not require manual input. Because sentiment information comes directly from real users, our method will adapt to user sentiment as the dialog evolves in real time. Another advantage of our model is that the sentiment scores come from a pre-trained sentiment detector, so no manual annotation of rewards is required. Furthermore, the sentiment information is independent of the user's goal, so no prior domain knowledge is required, which makes our method generalizable and independent of the task.


\section{Dataset}
\label{dataset}
We experimented our methods on DSTC1 dataset \cite{DSTC1}, which has a bus information search task. Although DSTC2 dataset is a more commonly-used dataset in evaluating dialog system performance, the audio recordings of DSTC2 are not publicly available and therefore, DSTC1 was chosen.  
There are a total of 914 dialogs in DSTC1 with both text and audio information. Statistics of this dataset are shown in Table \ref{table:statistics textdata}.  We used the automatic speech recognition (ASR) as the user text inputs instead of the transcripts, because the system's action decisions heavily depend on ASR. There are 212 system action templates in this dataset. Four types of entities are involved, \texttt{<place>}, \texttt{<time>}, \texttt{<route>}, and \texttt{<neighborhood>}. 

\section{Annotation}
\label{sec:annotation}
We manually annotated 50 dialogs consisting of 517 conversation turns for user sentiment. Sentiment is categorized into \texttt{negative, neutral} and \texttt{positive}. The annotator had access to the entire dialog history in the annotation process because the dialog context gives the annotators a holistic view of the interactions, and annotating user sentiment in a dialog without the context is really difficult. Some previous studies have also performed similar user information annotation given context, such as \citet{devillers2002annotation}.
The annotation scheme is described in Table \ref{table:emotion annotation} in Appendix \ref{appendix: annotation scheme}. To address the concern that dialog quality may bias the sentiment annotation, we explicitly asked the annotators to focus on users' behaviour instead of the system, and hid all the details of multimodal features from the annotators. Moreover, two annotators were calibrated on 37 audio files, and reached an inter-annotator agreement (kappa) of 0.74.  The statistics of the annotation results are shown in Table \ref{table:statistics audio}. The skewness in the dataset is due to the data's nature. In the annotation scheme, \texttt{positive} is defined as ``excitement or other positive feelings'', but people rarely express obvious excitement towards automated task-oriented dialog systems. What we really want to distinguish is neutral and positive cases from negative cases so as to avoid the negative sentiment, and the dataset is balanced for these two cases. To the best of our knowledge, our dataset is the first publicly available dataset that annotated user sentiment with respect to the entire dialog history. There are similar datasets with emotion annotations \cite{schuller2013interspeech} but are not labeled under dialog contexts.



\begin{table}
\centering
\resizebox{\columnwidth}{!}{
\begin{tabular}{cc}
\begin{tabular}{|l|l|}
\hline
{\bf Category } & {\bf Total}\\
\hline
total dialogs & 914 \\
total dialogs in train & 517 \\
total dialogs in test & 397 \\ 
\hline
\end{tabular} & 
\begin{tabular}{|l|l|}
\hline
{\bf Statistics} & {\bf  Total}\\\hline
avg dialog len & 13.8 \\ 
vocabulary size & 685 \\ 
\hline
\end{tabular}
\end{tabular}
}
\caption{Statistics of the text data.}
\label{table:statistics textdata}
\end{table}

\begin{table}
\centering
\resizebox{\columnwidth}{!}{
\begin{tabular}{cc}
\begin{tabular}{|l|l|}
\hline
{\bf  Category}         & {\bf Total}\\
\hline
total dialogs           & 50 \\
total audios            & 517 \\
total audios in train   & 318 \\ 
total audios in dev     & 99 \\ 
total audios in test    & 100 \\
\hline
\end{tabular} & 
\begin{tabular}{|l|l|}
\hline
{\bf Category}  & {\bf  Total}\\\hline
neutral         & 254 \\ 
negative        & 253 \\ 
positive        & 10 \\ 
\hline
\end{tabular}
\end{tabular}
}
\caption{Statistics of the annotated audio set.}
\label{table:statistics audio}
\end{table}

\section{Multimodal Sentiment Classification}
\label{detector}

To detect user sentiment, we extracted a set of acoustic, dialogic and textual features.

\subsection{Acoustic features}
\label{acoustic features}
We used openSMILE \cite{Eyben:2013:RDO:2502081.2502224} to extract acoustic features. Specifically, we used the paralinguistics  configuration from \citet{schuller2003hidden}, which includes 1584 acoustic features, such as pitch, volume and jitter. In order to avoid possible overfitting caused by the large number of acoustic features, we performed tree-based feature selection \cite{scikit-learn} to reduce the size of acoustic features to 20. The selected features are listed in Table \ref{table:selected features} in Appendix \ref{selected acoustic features}.

\subsection{Dialogic features}
Four categories of dialogic features are selected according to previous literature \cite{liscombe2005using} and the statistics observed in the dataset. We used not only the per-turn statistics of these features, but also the accumulated statistics of them throughout the entire conversation so that the sentiment classifier can also take the entire dialog context into consideration. 
\begin{description}
\itemsep-0.4em
\item[Interruption] is defined as the user interrupting the system speech. 
Interruptions occurred fairly frequently in our dataset (4896 times out of 14860 user utterances).


\item[Button usage] When the user is not satisfied with the ASR  performance of the system, he/she would rather choose to press a button for "yes/no" questions, so the usage of buttons can be an indication of negative sentiment. During DSTC1 data collection, users were notified about the option to use buttons, so this kind of information is available in the data.

\item[Repetitions] There are two kinds of repetitions: the user asks the system to repeat the previous sentence, and the system keeps asking the same question due to failures to catch some important entity. In our model, we combined these two situations as one feature because very few user repetitions occur in our data ($<$1\%). But for other data, it might be helpful to separate them.

\item[Start over] is active when the user chooses to restart the task in the middile of the conversation. The system is designed to give the user an option to start over after several turns. If the user takes this offer, he/she might have negative sentiment.
\end{description}
\subsection{Textual features}
We also noticed that the semantic content of the utterance was relevant to sentiment. So we used the entire dataset as a corpus and created a tf-idf vector 
for each utterance as textual features.

\subsection{Classification results}

The sentiment classifier was trained on the 50 dialogs annotated with sentiment labels. The predictions made by this classifier were used for the supervised learning and reinforcement learning in the later sections.
We used random forest as our classifier (an implementation from scikit-learn \cite{scikit-learn}), as we had limited annotated data. We separated the data to be 60\% for training, 20\% for validation and 20\% for testing. 
Due to the randomness in the experiments, we ran all the experiments 20 times and reported the average results of different models in Table \ref{table:results part 1}. We also conducted unpaired one-tailed t-test to assess the statistical significance.

We extracted 20 acoustic features, eight dialogic features  and 164 textual features.  From Table \ref{table:results part 1}, we see that the model combining all the three categories of features performed the best (0.686 in F-1, $p<1\mathrm{e}{-6}$ compared to acoustic baseline). One interesting observation is that by only using eight dialogic features, the model already achieved 0.596 in F-1. Another interesting observation is that using 164 textual features alone reached a comparable performance (0.664), but the combination of acoustic and textual features actually brought down the performance to 0.647. One possible reason is that the acoustic information has noise that confused the textual information when combined. But this observation doesn't necessarily apply to other datasets. The significance tests show that adding dialogic features improved the baseline significantly. For example, the model with both acoustic features and dialogic features are significantly better than the one with only acoustic features ($p < 1\mathrm{e}{-6}$). In Table \ref{table:user features}, we listed the dialogic features with their relative importance rank, which were obtained from ranking their feature importance scores in the classifier. We observe that ``total interruptions so far" is the most useful dialogic features to predict user sentiment. The sentiment detector trained will be integrated in the end-to-end learning described later.



\begin{table}
\small
\centering
\resizebox{0.8\columnwidth}{!}{
\begin{tabular}{|l|c|}
\hline
\textbf{Dialogic Features} & \textbf{\makecell{Relative Rank\\ of importance}}\\
\hline
total interruptions so far                & 	1\\
\hline
interruptions                    &     2\\
\hline
total button usages so far				   &  	 3\\
\hline
total repetitions so far   &	 4\\
\hline
repetition    &     5\\
\hline
button usage                 &     6\\
\hline
total start over so far &     7\\
\hline
start over			   &     8\\
\hline
\end{tabular}
}
\caption{Dialogic features' relative importance rank in sentiment detection.}
\label{table:user features}
\end{table}

\begin{table}[htb]
\centering
\resizebox{\columnwidth}{!}{
\begin{tabular}{|c|c|c|c|} 
\hline
    \textbf{Model} & \textbf{Avg. of F-1} & \textbf{Std. of F-1} & \textbf{Max of F-1} \\
\hline
          Acoustic features only   &  0.635 & 0.027 & 0.686 \\
\hline
          Dialogic features only   &   0.596 & 0.001 & 0.596 \\
          Textual features only $\ast$ & 0.664 & 0.010 & 0.685 \\
          Textual + Dialogic $\ast$ & 0.672 & 0.011 & 0.700 \\
          Acoustic + Dialogic $\ast$  &  0.680 & 0.019 & 0.707 \\
          Acoustic + Textual & 0.647 & 0.025 & 0.686 \\
          Acoustic + Dialogic + Text $\ast$   & \textbf{0.686} & \textbf{0.028} & \textbf{0.756}\\
\hline
\end{tabular}
}
\caption{Results of sentiment detectors using different features. The best result is highlighted in \textbf{bold} and * indicates statistical significance compared to the baseline, which is using acoustic features only. ($p<0.0001$)}
\label{table:results part 1}
\end{table}

\section{Supervised Learning (SL)}
\label{SL}


We incorporated the detected user sentiment from the previous section into a supervised learning framework for training end-to-end dialog systems. There are many studies on building a dialog system in a supervised learning setting (\citet{bordes2016learning,eric2017copy,seo2016query, liu2017end, li2017end, hybrid-code-networks-practical-efficient-end-end-dialog-control-supervised-reinforcement-learning-2}). Following these approaches, we treated the problem of dialog policy learning as a classification problem, which is to select actions among system action templates given conversation history. Specifically, we decided to adopt the framework of \textit{Hybrid Code Network (HCN)}  introduced in \citet{hybrid-code-networks-practical-efficient-end-end-dialog-control-supervised-reinforcement-learning-2}, because it is the current state-of-the-art model. We reimplemented \textit{HCN} and used it as the baseline system, given the absence of direct comparison on DSTC1 data. One caveat is that \textit{HCN} used action masks (bit vectors indicating allowed actions at certain dialog states) to prevent impossible system actions, but we didn't use hand-crafted action masks in the supervised learning setting because manually designing action masks for 212 action templates is very labor-intensive. This makes our method more general and adaptive to different tasks. All the dialog modules were trained together instead of separately. Therefore, our method is end-to-end trainable and doesn't require human expert involvement.

 We listed all the context features used in \citet{hybrid-code-networks-practical-efficient-end-end-dialog-control-supervised-reinforcement-learning-2} in Table \ref{table:context features in SL} in Appendix \ref{appendix: context features}. In our model, we added one more set of context features, the user-sentiment-related features. For entity extraction, given that the entity values in our dataset form a simple unique fixed set, we used simple string matching. We conducted three experiments: the first one used entity presences as context features, which serves as the baseline; the second one used entity presences in addition to all the raw dialogic features mentioned in Table \ref{table:user features}; the third experiment used the baseline features plus the predicted sentiment label by the pre-built sentiment detector (converted to one-hot vector) instead of the raw dialogic features. We used the entire DSTC1 dataset to train the supervised model. The input is the normalized natural language and the contexutal features, and the output is the action template id.
 We kept the same experiment setting in \citet{hybrid-code-networks-practical-efficient-end-end-dialog-control-supervised-reinforcement-learning-2}, e.g.  \texttt{last\_action\_taken} was also used as a feature, along with word embeddings \cite{mikolov2013efficient} and bag-of-words; LSTM with 128 hidden-units and AdaDelta optimizer \cite{zeiler2012adadelta} were used to train the model.


The results of different models are shown in Table \ref{table:results part2}. We observe that using the eight raw dialogic features did not improve turn-level F-1 score. One possible reason is that a total of eight dialogic features were added to the model, and some of them might contain noises and therefore caused the model to overfit. However, using predicted sentiment information as an extra feature, which is a more condensed information, outperformed the other models both in terms of turn-level F-1 score and dialog accuracy which indicates if all turns in a dialog are correct.  The difference in absolute F-1 score is small because we have a relatively large test set (5876 turns). But the unpaired one-tailed t-test shows that $p<0.01$ for both the F-1 and the dialog accuracy.  This suggests that including user sentiment information in action planning is helpful in a supervised learning setting.


\begin{table}[h!]
\centering
\resizebox{\columnwidth}{!}{
\begin{tabular}{|l|c|c|} 
\hline
    \textbf{Model} & \textbf{Weighted F-1} & \textbf{Dialog Acc.} \\ [0.5ex] 
\hline
          HCN   &    0.4198  & 6.05\%  \\ 
\hline
          HCN + raw dialogic features   &   0.4190 &  5.79\%  \\
\hline
          HCN + predicted sentiment label$\ast$   &    \textbf{0.4261}  & \textbf{6.55\%}  \\
       \hline
\end{tabular}
}
\caption{Results of different SL models. The best result is highlighted in \textbf{bold}. $\ast$ indicates that the result is significantly better than the baseline ($p < 0.01$). Dialog accuracy indicates if all turns in a dialog are correct, so it's low. For DSTC2 data, the state-of-art dialog accuracy is 1.9\%, consistent with our results.  }
\label{table:results part2}
\end{table}

\section{Reinforcement Learning (RL)}
\label{RL}
In the previous section, we discussed including sentiment features directly as a context feature in a supervised learning model for end-to-end dialog system training, which showed promising results. But once a system
operates at scale and interacts with a large number
of users, it is desirable for the system to continue
to learn autonomously using reinforcement learning
(RL). With RL, each turn receives a measurement of goodness called \textit{reward} \cite{hybrid-code-networks-practical-efficient-end-end-dialog-control-supervised-reinforcement-learning-2}. Previously, training task-oriented systems mainly relies on the delayed reward about task success. Due to the lack of informative immediate reward, the training takes a long time to converge. In this work, we propose to include user sentiment as immediate rewards to expedite the reinforcement learning training process and create a better user experience.

To use sentiment scores in the reward function, we chose the \textit{policy gradient} approach \cite{williams1992simple} and implemented the algorithm based on \citet{zhu2017github}. The traditional reward function uses a positive constant (e.g. 20) to reward the success of the task, 0 or a negative constant to penalize the failure of the task after certain number of turns, and gives -1 to each extra turn to encourage the system to complete the task sooner. However, such reward function doesn't consider any feedback from the end-user. It is natural for human to consider conversational partner's sentiment in planning dialogs. So, we propose a set of new reward functions that incorporate user sentiment to emulate human behaviors.

The intuition of integrating sentiment in reward functions is as follows. The ultimate evaluator of dialog systems is the end-users. And user sentiment is a direct reflection of user satisfaction. Therefore, we detected the user sentiment scores from multimodal sources on the fly, and used them as immediate rewards in an adaptive end-to-end dialog training setting. This sentiment information came directly from real users, which made the system adapt to individual user's sentiment as the dialog proceeds. Furthermore, the sentiment information is independent of the task, so our method doesn't require any prior domain knowledge and can be easily generalized to other domains. There have been works that incorporated user information into reward design \cite{su2015learning, ultes2017domain}. But they used information from one single channel and sometimes required manual labelling of the reward. Our approach utilizes information from multiple channels and doesn't involve manual work once a sentiment detector is ready. 

We built a simulated system in the same bus information search domain to test the effectiveness of using sentiment scores in the reward function.  In this system, there are 3 entity types - \texttt{<departure>, <arrival>}, and \texttt{<time>} - and 5 actions, asking for different entities, and giving information. A simple action mask was used to prevent impossible actions, such as giving information of an uncovered place. The inputs to the system are the simulated user's dialog acts and the simulated sentiment sampled from a subset of DSTC1, the $CleanData$, which will be described later. The output of the system is the system action template.

\subsection{User simulator}
Given that reinforcement learning requires feedback from the environment - in our case, the users - and interacting with real users is always expensive, we created a user simulator to interact with the system. At the beginning of each dialog, the simulated user is initiated with a goal consisting of the three entities mentioned above and the goal remains unchanged throughout the  conversation. The user responds to system's questions with entities, which are placeholders like \texttt{<departure>} instead of real values. To simulate ASR errors, the simulated user's act type occasionally changes from ``informing slot values'' to ``making noises'' at certain probabilities set by hand (10\% in our case). 
Some example dialogs along with their associated rewards are shown in Table \ref{table:RL baseline} and \ref{table:RL penalty 2} in Appendix \ref{sec:reward example}.

We simulated user sentiment by sampling from real data, the DSTC1 dataset. There are three steps involved. First, we cleaned the DSTC1 dialogs by removing the audio files with no ASR output and high ASR errors. This resulted in a dataset $CleanData$ with 413 dialogs and 1918 user inputs. We observed that users accumulate their sentiment as the conversation unfolds. When the system repeatedly asks for the same entity, they express stronger sentiment. Therefore, summary statistics that record how many times certain entities have been asked during the conversation is representative of users' accumulating sentiment. We designed a set of summary statistics $S$ that record the statistics of system actions, e.g. how many times the arrival place has been asked or the schedule information has been given. 

The second step is to create a mapping between the five simulated system actions and the DSTC1 system actions. We do this by calculating a vector $s_{real}$ consisting of the values in $S$ for each user utterance in $CleanData$. $s_{real}$ is used to compare the similarity between the real dialog and the simulated dialog.

The final step is to sample from $CleanData$. For each simulated user utterance, we calculated the same vector $s_{sim}$ and compared it with each $s_{real}$. There are two possible results. If there are $s_{real}$ equal to $s_{sim}$,
we would randomly sample one from all the matched user utterances to represent the sentiment of the simulated user. But if there is no  $s_{real}$ matching $s_{sim}$, different strategies would be applied based on the reward function used, which will be described in details later. Once we have a sample, the eight dialogic features of the sample utterance are used to calculate the sentiment score. We didn't use the acoustic or the textual features because in a simulated setting, only the dialogic features are valid.


\subsection{Experiments}
We designed four experiments with different reward functions. A discount factor of 0.9 was applied to all the experiments. And the maximum number of turns is 15. Following \citet{hybrid-code-networks-practical-efficient-end-end-dialog-control-supervised-reinforcement-learning-2}, we used LSTM with 32 hidden units for the RNN in the \textit{HCN} and AdaDelta for the optimization, and updated the reinforcement learning policy after each dialog. The $\epsilon$-greedy exploration strategy \cite{tokic2010adaptive} was applied here. Given that the entire system was simulated, we only used the presence of each entity and the last action taken by the system as the context features, and didn't use bag-of-words or utterance embedding features.

In order to evaluate the method, we froze the policy after every 200 updates, and ran 500 simulated dialogs to calculate the task success rate. We repeated the process 20 times and reported the average performance in Figure \ref{fig:RL len}, \ref{fig:RL success} and Table \ref{table:RL success other models}.

\subsubsection{Baseline}

We define the baseline reward as follows without any sentiment involvement.

\floatname{algorithm}{Reward}
\begin{algorithm}[H]
\scriptsize
\caption{Baseline}
\begin{algorithmic} 
\IF{success}
	\STATE $R_1 = 20 $
\ELSIF{failure}
	\STATE $R_1 = -10 $
\ELSIF{each proceeding turn}
	\STATE $R_1 = -1 $
\ENDIF
\end{algorithmic}
\end{algorithm}

\subsubsection{Sentiment reward with random samples (SRRS)}
We designed the first simple reward function with user sentiment as the immediate reward: sentiment with random samples (SRRS). We first drew a sample from real data with matched context; if there was no matched data, a random sample was used instead. Because the amount of $CleanData$ is relatively small, so only 36\% turns were covered by matched samples.
If the sampled dialogic features were not all zeros, the sentiment reward ($SR$) was calculated as a linear combination with tunable parameters. We chose it to be \texttt{-5P\textsubscript{neg}-P\textsubscript{neu}+10P\textsubscript{pos}} for simplicity.
When the dialogic features were all zero, in most cases it meant the user didn't express an obvious sentiment, we set the reward to be -1.

\begin{algorithm}[H]
\scriptsize
\caption{SRRS}
\begin{algorithmic} 
\IF{success}
	\STATE $R_2 = 20 $
\ELSIF{failure}
	\STATE $R_2 = -10 $
\ELSIF{sample with all-zero dialogic features}
	\STATE $R_2 = -1 $
\ELSIF{sample with non-zero dialogic features}
	\STATE $R_2\text{=-5P\textsubscript{neg}-P\textsubscript{neu}+10P\textsubscript{pos}}$
\ENDIF
\end{algorithmic}
\end{algorithm}


\subsubsection{Sentiment reward with repetition penalty (SRRP)}
Random samples in SRRS may result in extreme sentiment data. So we used dialogic features to approximate sentiment for the unmatched data.  Specifically, if there were repetitions, which correlate with negative sentiment (see Table \ref{table:user features}), we assigned a penalty to that utterance. See Reward 3 Formula below for detailed parameters. 36\% turns were covered by real data samples, 15\% turns had no match in real data and had repetitions, and 33\% turns had no match and no repetition. 

Moreover, we experimented with different penalty weights. When we increased the repetition penalty to 5, the success rate was similar to penalty of 2.5. However, when we increased the penalty even further to 10, the success rate was brought down by a large margin. Our interpretation is that increasing the repetition penalty to a big value made the focus less on the real sentiment samples but more on the repetitions, which did not help the learning.

\begin{algorithm}[H]
\scriptsize
\caption{SRRP}
\begin{algorithmic} 
\IF{success}
	\STATE $R_3 = 20 $
\ELSIF{failure}
	\STATE $R_3 = -10 $
\ELSE
    \IF{match}
        \IF{all-zero dialogic features}
        	\STATE $R_3 = -1 $
        \ELSIF{non-zero dialogic features}
        	\STATE $R_3\text{=-5P\textsubscript{neg}-P\textsubscript{neu}+10P\textsubscript{pos}}$
        \ENDIF
    \ELSIF{repeated question}
        \STATE $R_3 = -2.5 $ 
    \ELSE
        \STATE $R_3 = -1 $
    \ENDIF
\ENDIF
\end{algorithmic}
\end{algorithm}

%

\subsubsection{Sentiment reward with repetition and interruption penalties (SRRIP)}
We observed in Section \ref{detector} that \texttt{interruption} is the most important feature in detecting sentiment, so if an interruption existed in the simulated user input, we assumed it had a negative sentiment and added an additional penalty of -1 to the previous sentiment reward SRRP to test the effect of interruption.  7.5\% turns have interruptions. 


\begin{algorithm}[H]
\scriptsize
\caption{SRRIP}
\begin{algorithmic} 
\IF{success}
	\STATE $R_4 = 20 $
\ELSIF{failure}
	\STATE $R_4 = -10 $
\ELSE
    \IF{match}
        \IF{all-zero dialogic features}
        	\STATE $R_4 = -1 $
        \ELSIF{non-zero dialogic features}
        	\STATE $R_4\text{=-5P\textsubscript{neg}-P\textsubscript{neu}+10P\textsubscript{pos}}$
        \ENDIF
    \ELSIF{repeated question}
        \STATE $R_4 = -2.5 $ 
    \ELSE
        \STATE $R_4 = -1 $
    \ENDIF
    \IF{interruption}
        \STATE $R_4 = R_4 - 1$
    \ENDIF
\ENDIF
\end{algorithmic}
\end{algorithm}




\subsection{Experiment results}

\begin{figure}[h]
  \includegraphics[width=\columnwidth, height=6cm]{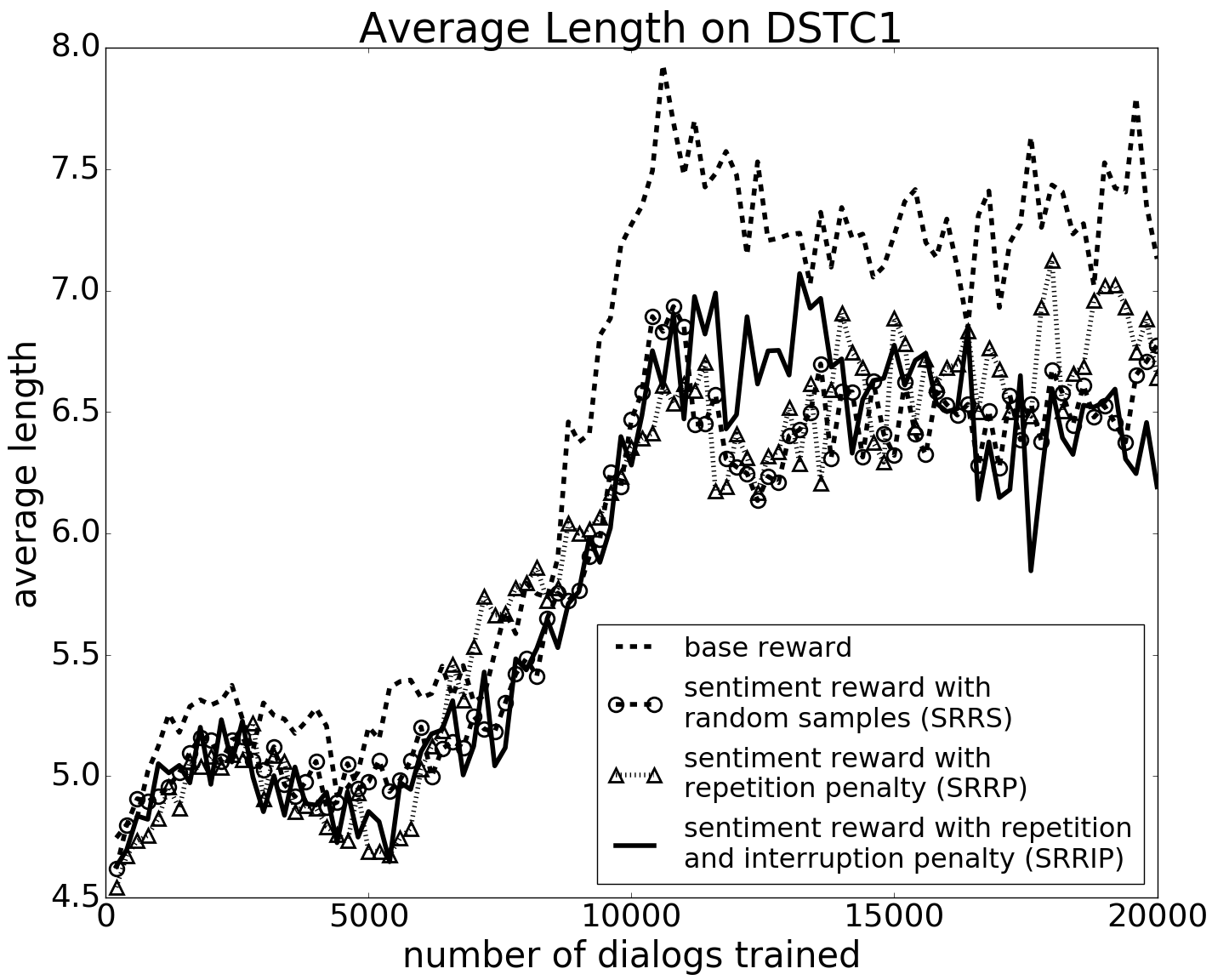}
  \caption{Average dialog length of RL models with different reward functions.}
  \label{fig:RL len}
\end{figure}

We evaluated every model on two metrics: dialog lengths and task success rates. 
We observed in Figure \ref{fig:RL len} that all the sentiment reward functions, even SRRS with random samples, reduced the average length of the dialogs, meaning that the system finished the task faster. The rationale behind is that by adapting to user sentiment, the model can avoid unnecessary system actions to make systems more effective. 

\begin{figure}[h]
  \includegraphics[width=\columnwidth, height=6cm]{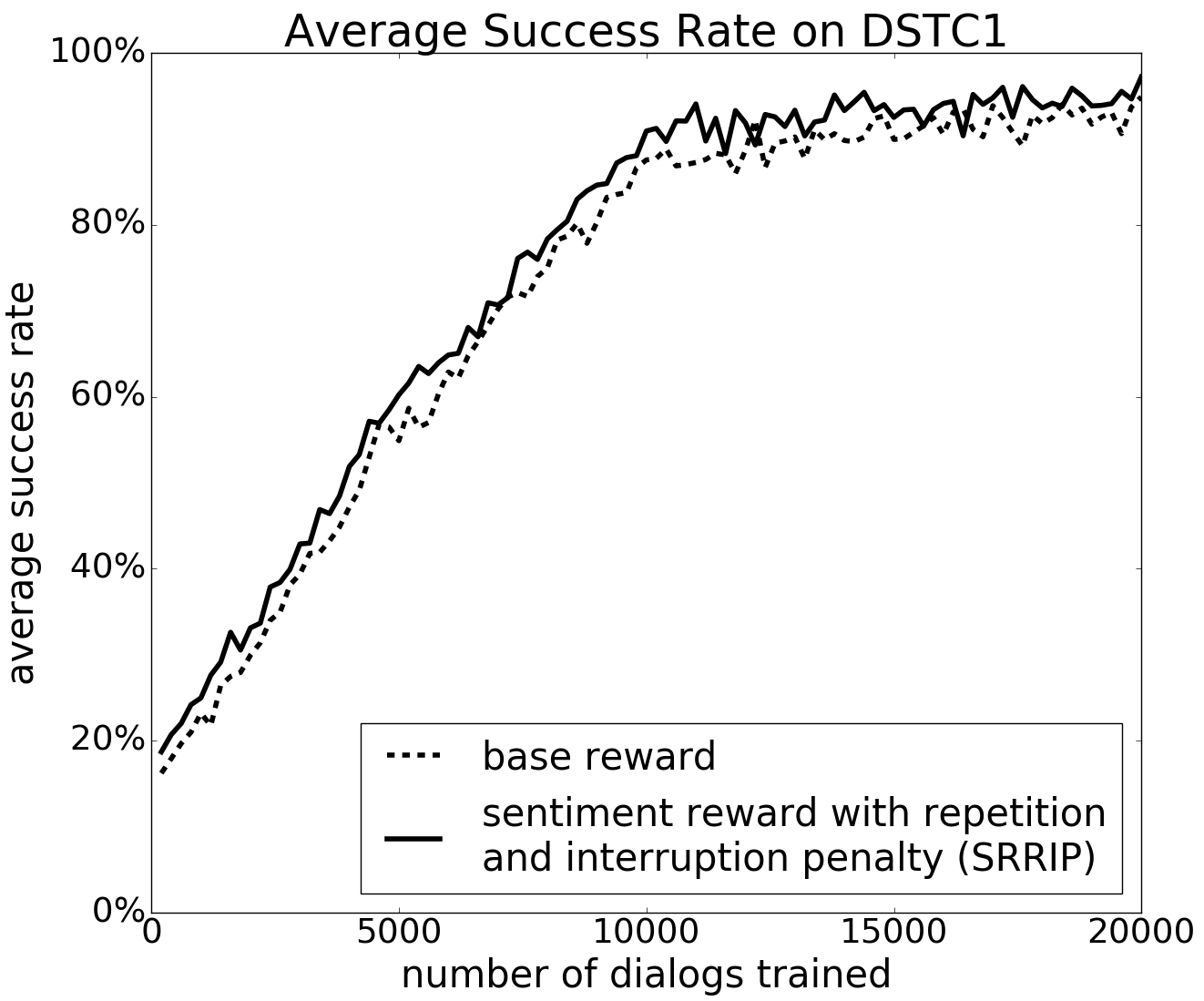}
  \caption{Average success rate of the baseline and the best performing model, SRRIP.}
  \label{fig:RL success}
\end{figure}

In terms of success rate, sentiment reward with both repetition and interruption penalties (SRRIP) performed the best (see Figure \ref{fig:RL success}). 
In Figure \ref{fig:RL success}, SRRIP is converging faster than the baseline. For example, around 5000 iterations, it outperforms the baseline by 5\% in task success rate (60\% vs 55\%) with statistical significance ($p < 0.01$). It also converges to a better task success rate after 10000 iterations (92.4\% vs 94.3\%, $p<0.01$). 

We describe all models' performance in Table \ref{table:RL success other models} in terms of 
the convergent success rate calculated as the mean success rate after 10000 dialogs. 
We observed that incorporating various sentiment rewards improved the success rate and expedited the training process overall with statistical significance. We found that even sentiment reward with random samples (SRRS) outperformed the baseline after convergence. By adding penalties for repetition, the algorithm covered more data points, and therefore, the task success rate and the convergence speed improved. We also found that penalizing interruption and repetition together (SRRIP) achieved a slightly better performance compared to penalizing repetition only (SRRP). This suggests that interruptions is another factor to consider when approximating sentiment. But the performances between SRRP and SRRIP is not significant. Our guess is that only 7.5\% turns in our data contains interruption and the penalty is just an extra -1, so the model confused this signal with noises. But given more interruptions in the data, interruptions could still be helpful. 



\begin{table}
\centering
\begin{tabular}{|c|c|} 
\hline
    \textbf{Model} &  \textbf{\makecell{Convergent \\success rate}} \\ [0.5ex] 
\hline
          Baseline   &     0.924  \\ 
\hline
          SRRS   &         0.938$\ast$  \\
\hline
          SRRP    &         0.941$\ast$  \\
\hline
        \textbf{SRRIP} &    \textbf{0.943$\ast$}\\
\hline
\end{tabular}
\caption{Convergent success rate of RL models with different reward functions. It is calculated as the mean success rate after 10000 dialogs.  The best result is highlighted in \textbf{bold}. $\ast$ indicates that the result is significantly better than the baseline ($p < 0.01$).}
\label{table:RL success other models}
\end{table}


\section{Discussion and Future Work}
The intuition behind the good performance of models with user sentiment is that the learned policy is in general more sentiment adaptive.  For example, there are some system actions that have the same intention but with different surface forms, especially for error-handling strategies. By analyzing the results, we found that when the sentiment adaptive system detected a negative sentiment from the user, it chose to respond the user with a more detailed error-handling strategy than a general one. For example, it chose the template ``Where are you leaving from? For example, you can say,  $<$place$>$'', while the baseline model would respond with ``Where would you like to leave from?'', which doesn't provide details to help the user compared with the previous template. As we all know, dealing with a disappointed user to proceed, providing more details is always better. One example dialog is shown in Table \ref{table:personal example}. There was no written rules to force the model to choose one specific template under certain situations, so the model learned these subtle differences on its own. Some may argue that the system could always use a more detailed template to better guide the user instead of distinguishing between two similar system templates. But this is not necessarily true. Ideally, we want the system to be succinct initially to save users' time, because we observe that users, especially repeated users, tend to interrupt long and detailed system utterances. If the user has attempted to answer the system question but failed, then it's beneficial to provide detailed guidance.




\begin{table}
\small
\begin{tabularx}{\columnwidth}{|X|X|} 
\hline
   \textbf{Sentiment Adaptive System} & \textbf{Baseline System without Sentiment}\\
\hline
\textbf{SYS}: The \texttt{<route>}. Where would you like to leave from? & \textbf{SYS}: The \texttt{<route>}. Where would you like to leave from? \\
\hline
\textbf{USR}: \textit{Yeah \textbf{[negative sentiment]}} & \textbf{USR}: \textit{Yeah} \\
\hline
\textbf{SYS}: Where are you leaving from? For example, you can say, \texttt{<place>}. & \textbf{SYS}: Right. Where would you like to leave from?\\
\hline
\end{tabularx}
\caption{An example dialog by different systems in the supervised learning setting. The sentiment-adaptive system gives a more detailed error-handling strategy than the baseline system.}
\label{table:personal example}
\end{table}


The performance of the sentiment detector is a key factor in our work. So in the future, we plan to incorporate features from more channels such as vision to further improve the sentiment predictor's performance, and potentially further improve the performance of the dialog system. We also want to explore more in user sentiment simulation, for example, instead of randomly sampling data for the uncovered cases, we could use linear interpolation to create a similarity score between $s_{sim}$ and $s_{real}$, and choose the user utterance with the highest score. Furthermore, reward shaping \cite{ng1999policy, ferreira2013expert} is an important technique in RL. Specifically, \citet{ferreira2013expert} talked about incorporating expert knowledge in reward design. We also plan to integrate information from different sources into reward function and apply reward shaping. 
Besides, creating a good user simulator is also very important in the RL training. There are some more advanced methods to create user simulators. For example, \citet{liu2017iterative} described how to optimize the agent and the user simulators jointly using RL. We plan to apply our sentiment reward functions in this framework in the future.

\section{Conclusion}
\label{Result}
We proposed to detect user sentiment  from multimodal channels and incorporate the detected sentiment as feedback into adaptive end-to-end dialog system training to make the system more effective and user-adaptive. We included sentiment information directly as a context feature in the supervised learning framework and used sentiment scores as immediate rewards in the reinforcement learning setting. Experiments suggest that incorporating user sentiment is helpful in reducing the dialog length and increasing the task success rate in both SL and RL settings. This work proposed an adaptive methodology to incorporate user sentiment in end-to-end dialog policy learning and showed promising results on a bus information search task. We believe this approach can be easily generalized to other domains given its end-to-end training procedure and task independence.


\section*{Acknowledgments}

The work is partly supported by Intel Lab Research Gift.

\bibliography{acl2018}
\bibliographystyle{acl_natbib}

\appendix

\section{Supplemental Material}
\label{sec:supplemental}

\subsection{Example dialogs with different reward functions}
\label{sec:reward example}

\begin{table}[H]
\small
\centering
\begin{tabularx}{0.95\columnwidth}{|>{\hsize=0.7\hsize}X|
                        >{\hsize=0.19\hsize}X|
                        >{\hsize=0.11\hsize}X|} 
\hline
   \textbf{Dialogs} & \multicolumn{1}{|l|}{\textbf{Category}} & \multicolumn{1}{|l|}{\textbf{Reward}}  \\ 
\hline
\textbf{USR}: \textit{I am at \texttt{<uncovered\_departure>}.} & &   \\ 
\hline
\textbf{SYS}: what time do you want to travel? & proceeding turn &-1 \\
\hline
\textbf{USR}: \textit{At \texttt{<time>}.}       		  & &   \\ 
\hline
\textbf{SYS}: what time do you want to travel? & proceeding turn &-1 \\
\hline
\textbf{USR}: \textit{You already knew that!}      	  &  &  \\ 
\hline
\textbf{SYS}: Where do you want to go?      	  & proceeding turn &-1 \\ 
\hline
\textbf{USR}: \textit{I want to go to \texttt{<uncovered\_arrival>}.} & &   \\ 
\hline
\textbf{SYS}: Let me look that up for you. Sorry, there is no result that matches your request. & success & 20 \\
\hline
   & \textbf{Total reward} & 17 \\
\hline
\end{tabularx}
\caption{An example of the simulated dialog using the baseline reward.}
\label{table:RL baseline}
\end{table}

\begin{table}[H]
\small
\begin{tabularx}{\columnwidth}{|>{\hsize=0.7\hsize}X|
                        >{\hsize=0.17\hsize}X|
                        >{\hsize=0.13\hsize}X|} 
\hline
    \textbf{Dialogs} & \multicolumn{1}{|l|}{\textbf{Category}} & \multicolumn{1}{|l|}{\textbf{Reward}}  \\ 
\hline
\textbf{USR}: \textit{I am at \texttt{<uncovered\_departure>}} & &    \\ 
\hline
\textbf{SYS}: Where do you want to go? & no match & -1 \\
\hline
\textbf{USR}: \textit{\texttt{<arrival>}.}       		  & &   \\ 
\hline
\textbf{SYS}: where do you want to go? & matched & -4.01 \\
\hline
\textbf{USR}:\textit{ You already knew that!}       		  & &   \\ 
\hline
\textbf{SYS}: where are you leaving from? & repetition & -2.5 \\
\hline
\textbf{USR}: \textit{You already knew that!}    	       & &  \\ 
\hline
\textbf{SYS}: what time do you want to travel?  & no match & -1 \\ 
\hline
\textbf{USR}: \textit{At \texttt{<time>}.} &       &               \\ 
\hline
\textbf{SYS}: Let me look that up for you. Sorry, there is no result that matches your request.     & success &20 \\
\hline
& \textbf{Total reward} & 11.49 \\
       \hline
\end{tabularx}
\caption{An example of the simulated dialog using the sentiment reward with repetition penalty (SRRP).}
\label{table:RL penalty 2}
\end{table}

\subsection{Annotation scheme}
\label{appendix: annotation scheme}

\begin{table}[H]
\small
\centering
\begin{tabularx}{0.95\columnwidth}{|>{\hsize=0.1\hsize}X|
                        >{\hsize=0.17\hsize}X|
                        >{\hsize=0.73\hsize}X|}
\hline 
\multicolumn{1}{|l|}{\bf Label} & \multicolumn{1}{|l|}{\bf Definition} & \bf Description \\ 
\hline
-1 & negative & Shows impatience, disappointment, anger or other negative feelings in voice. \\
\hline
0 & neutral & Shows neither positive nor negative feelings in voice. \\
\hline
1 & positive & Shows excitement, happiness or other positive feelings in voice.\\
\hline
\end{tabularx}
\caption{Sentiment annotation scheme.}
\label{table:emotion annotation}
\end{table}

\subsection{Context features used in supervised learning}
\label{appendix: context features}

\begin{table}[H]
\small
\centering
\begin{tabularx}{0.9\columnwidth}{|>{\hsize=0.35\hsize}X|
                        >{\hsize=0.65\hsize}X|} 
\hline
    \textbf{Model} & \textbf{Context features}   \\ [0.5ex] 
\hline
    baseline HCN   & Presence of each entity in dialog state    \\ 
\hline
    HCN + dialogic features & Presence of each entity + dialogic features in Table \ref{table:user features}   \\
\hline
    HCN + predicted sentiment label  & Presence of each entity + predicted sentiment label in one-hot vector   \\
       \hline
\end{tabularx}
\caption{Context features in different SL models.}
\label{table:context features in SL}
\end{table}

\subsection{Selected acoustic features in Section \ref{acoustic features}}
\label{selected acoustic features}
\begin{table}[H]
\centering
\resizebox{0.8\columnwidth}{!}{
\begin{tabular}{|l|}
\hline
\textbf{Selected acoustic features after feature selection} \\
\hline
pcm\_loudness\_sma\_iqr1-3 \\
pcm\_fftMag\_mfcc\_sma[0]\_pctlrange0-1 \\
pcm\_fftMag\_mfcc\_sma[0]\_upleveltime90 \\
pcm\_fftMag\_mfcc\_sma[7]\_upleveltime75 \\
logMelFreqBand\_sma[1]\_quartile3 \\
logMelFreqBand\_sma[1]\_upleveltime75 \\
logMelFreqBand\_sma[2]\_kurtosis \\
logMelFreqBand\_sma[3]\_amean \\
logMelFreqBand\_sma[3]\_linregc2 \\
logMelFreqBand\_sma[5]\_upleveltime90 \\
lspFreq\_sma[7]\_minPos \\
lspFreq\_sma[7]\_skewness \\
pcm\_loudness\_sma\_de\_kurtosis \\
pcm\_fftMag\_mfcc\_sma\_de[5]\_amean \\
pcm\_fftMag\_mfcc\_sma\_de[6]\_linregerrQ \\
pcm\_fftMag\_mfcc\_sma\_de[9]\_kurtosis \\
pcm\_fftMag\_mfcc\_sma\_de[10]\_iqr1-2 \\
logMelFreqBand\_sma\_de[5]\_quartile2 \\
lspFreq\_sma\_de[4]\_percentile99.0 \\
jitterLocal\_sma\_de\_iqr1-2 \\
\hline
\end{tabular}
}
\caption{Selected acoustic features in Section \ref{acoustic features}}
\label{table:selected features}
\end{table}

\end{document}